\documentclass[letterpaper, 10 pt, conference]{ieeeconf}

\IEEEoverridecommandlockouts
\usepackage{amsthm}
\newtheorem{theorem}{Theorem}
\newtheorem{proposition}{Proposition}

\usepackage{cite}
\usepackage{graphicx}
\usepackage{subcaption}
\usepackage{amsmath,amssymb,amsfonts}
\usepackage{bm}
\usepackage{xcolor}
\usepackage{gensymb}
\usepackage{siunitx}
\usepackage{booktabs}
\usepackage{multirow}

\newtheorem{remark}{Remark}

\begin{document}

\title{\LARGE \bf
Force-Closure Condition for Cooperative\\
Multi-robot Pushing of Rectangular Objects
}



\author{Duy Anh Nguyen$^{1}$ 
    and Trung Dung Ngo$^{1*}$
\thanks{We acknowledge the support of the Natural Sciences and Engineering Council of Canada, Department of National Defence: NSERC-RGPIN-2023-05961, DGDND-2023-05961, NSERC-CREATE 528099-2019, and NSERC-Alliance ALLRP 593651-2024. } 
\thanks{$^{1}$Duy Anh Nguyen and Trung Dung Ngo are with the More-Than-One Robotics Laboratory, Faculty of Sustainable Design Engineering, University of Prince Edward Island, Charlottetown, Canada.}%
\thanks{$^{*}$Corresponding author. Email: {\tt\small  tngo@upei.ca}}%
}

\maketitle

\begingroup
\renewcommand{\thefootnote}{}
\footnotetext{%
\scriptsize
\textcopyright{} 2026 IEEE. Personal use of this material is permitted.
Permission from IEEE must be obtained for all other uses, in any current or future media,
including reprinting/republishing this material for advertising or promotional purposes,
creating new collective works, for resale or redistribution to servers or lists,
or reuse of any copyrighted component of this work in other works.
\par\smallskip
Published in 2026 IEEE Canadian
Atlantic Ocean Symposium (CAOS), pp. 165--170.
DOI: 10.1109/CAOS69069.2026.11665860.
}
\endgroup

\begin{abstract}
In cooperative object manipulation, the placement of pushing contacts determines whether force closure can be achieved and agents can generate arbitrary forces and torques on the object correspondingly. This study derives a necessary and sufficient closed-form condition for the planar force closure of a rectangular object, a common object shape in everyday settings, under four frictionless pushing contacts. For a rectangle of length $b$ and width $a$, force closure holds if and only if $(l_1+l_3-b)(l_2+l_4-a)<0$, where $l_1,l_2,l_3,l_4$ are the four edge contact locations. This condition is independent of the object's center of mass and is used to verify contact configurations before contact force allocation. When the condition is not satisfied, the signs of the two factors indicate which opposing contact pair should be adjusted. The condition and adjustment rule are then integrated into a full-pose trajectory-tracking framework as a geometric feasibility layer to demonstrate their use in control. Static wrench feasibility tests validate the proposed condition, while narrow-passage floating-body simulations show that maintaining a force-closure configuration reduces the RMS tracking errors in $x$, $y$, and unwrapped yaw from $12.05$~m, $18.11$~m, and $535.96^\circ$ to $2.24$~m, $2.66$~m, and $2.69^\circ$, respectively.
\end{abstract}


\section{INTRODUCTION}

\begin{figure}[!t]
    \centering
    \vspace{0.6em}
    \includegraphics[
        width=0.75\linewidth,
        trim={0 20 0 25},
        clip
    ]{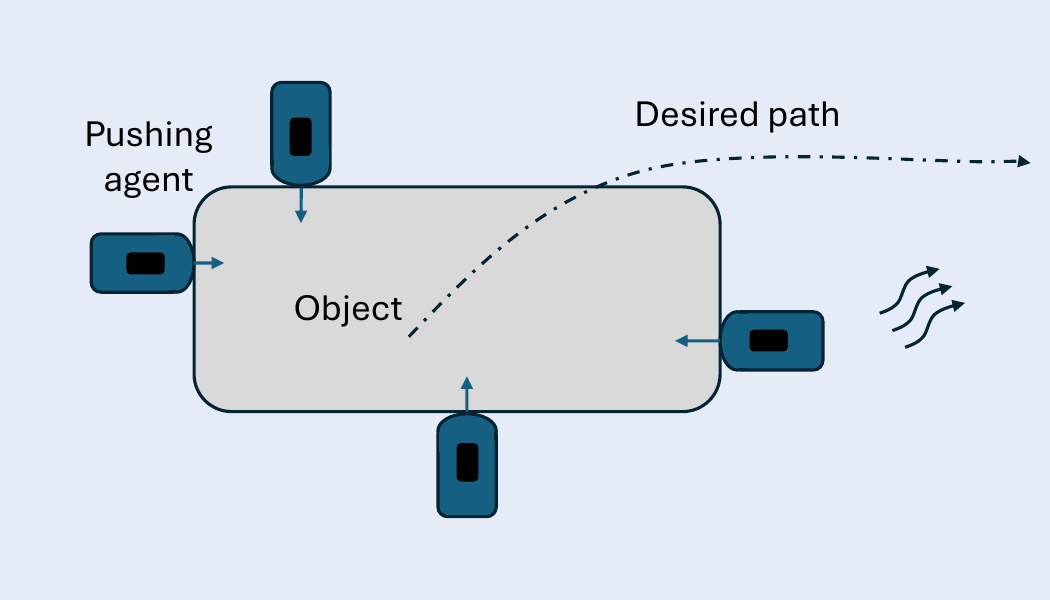}
    \caption{Cooperative pushing of a rectangular object by multiple agents, motivated by marine applications such as tug-assisted floating body maneuvering. Suitable contact placement is required to generate the planar force and yaw moment needed for object motion control.}
    \label{fig_intro}
    \vspace{-2mm}
\end{figure}

Cooperative object manipulation by multiple robots becomes necessary when a task exceeds the force, moment, or workspace capacity of any single agent~\cite{Review_Tuci,du2023review}. In marine settings, tugboats are deployed in groups to dock large vessels, reposition heavy floating barges, and move floating structures through narrow waterways, as illustrated in Fig.~\ref{fig_intro}. Multiple agents, such as robots or vessels, can manipulate a shared floating object in several ways, including attaching, caging, towing, and pushing~\cite{du2023review}. Among these, pushing requires the least mechanical setup: it needs no cables, attachments, or grasping mechanisms, and tugboats carry it out by pressing their bow or stern fenders against the hull of the target object.

Cooperative pushing is nontrivial because object motion depends on both the control law and the contact locations. Under unilateral frictionless pushing, the contact normals and nonnegative contact force magnitudes define a wrench cone; if the commanded object wrench lies outside this cone, no nonnegative contact force can realize it~\cite{murray1994mathematical}.  Force closure provides the classical framework for this feasibility check. A contact configuration achieves force closure when contact forces can balance any external wrench on the object, or equivalently generate arbitrary planar wrenches \cite{murray1994mathematical}. Nguyen's work established a constructive approach for synthesizing planar force closure grasps from the geometry of the object \cite{nguyen1988constructing}. The method identifies independent contact regions and provides algorithms for constructing grasps from object shape information. These works provide a basis for contact-feasibility analysis, but rely on general geometric or algorithmic frameworks rather than a compact closed-form condition for the cooperative pushing geometry considered here.

This contact feasibility issue is especially relevant in cooperative object manipulation, where multiple agents must jointly generate the forces and moments required for transport. Esposito~\cite{esposito2008distributed} studied swarm-based cooperative pushing of floating objects in autonomous tugboat scenarios, highlighting how robot placement affects the manipulation capability of the team. Cooperative floating object transport has also been studied through layered control frameworks involving trajectory tracking, wrench allocation, and vessel formation control \cite{chen2019distributed}. More generally, constrained control allocation has been widely studied for marine vehicles \cite{fossen2006survey,johansen2013control}. Autonomous tugboat systems have addressed towing, waypoint following, heading regulation, and collision avoidance~\cite{du2020cooperative,du2022multi}. Related pushing-based work with soft tactile sensing handled distributed force regulation and motion direction control for a rectangular object~\cite{SII}, where feasibility is assessed through moment arm signs relative to the object center of mass for a specific target direction. Recent work has also addressed collaborative planar pushing of polytopic objects using hybrid mode optimization and nonlinear MPC~\cite{tang2024collaborative}. These approaches either assume that the selected contact geometry can produce the required object wrench or evaluate feasibility through numerical optimization. For one unilateral edge normal contact on each edge of a rectangular body, a closed-form condition that is independent of the object's center of mass remains underexplored.

This study tackles these gaps by deriving a necessary and sufficient closed-form condition for force closure with four frictionless, unilateral contacts on a rectangular floating object, stated purely in terms of the four edge contact positions. The condition is a single checkable algebraic inequality that requires no wrench-space computation and can be evaluated at the contact placement stage before applying a control technique. To demonstrate its use in control, we embed this result into a simple 3-DOF floating-body trajectory-tracking framework: the condition is checked once upfront to confirm that the contact geometry supports full-pose tracking, and if not, a pair-and-direction adjustment rule corrects the configuration before contact force allocation. In this framework, the pushing agents are modeled as ideal unilateral normal-force actuators at the contact points, allowing the study to isolate the effect of contact geometry on wrench generation. Simulation results demonstrate that contact configurations satisfying the proposed force-closure condition can produce the commanded planar forces and yaw moments required for full-pose tracking, while configurations that violate the condition lead to wrench errors and poorer tracking performance.

The paper is organized as follows: Section~II covers the problem formulation, Section~III derives the force-closure condition, Section~IV presents the tracking framework, Section~V reports simulation results, and Section~VI concludes.

\section{PROBLEM FORMULATION}
\subsection{Object model}

Consider a floating rectangular object moving on a horizontal plane, 
modeled as a reduced 3-DOF planar floating body following the standard 
marine craft formulation~\cite{fossen2011handbook}. Let the object pose 
in the world frame and velocity in the body frame be:
\begin{equation}
\eta = \begin{bmatrix} x & y & \psi \end{bmatrix}^\top,\qquad
\nu = \begin{bmatrix} u & v & r \end{bmatrix}^\top,
\end{equation}
where $(x,y)$ denotes the object position, $\psi$ is the yaw angle, 
and $(u,v,r)$ are the surge, sway, and yaw velocities, respectively. 
The equations of motion are:
\begin{equation}
\dot{\eta} = J(\psi)\nu, \qquad M\dot{\nu} + D\nu = \tau_c,
\end{equation}
where $J(\psi)$ is the rotation matrix from body to world frame,
\begin{equation}
J(\psi)=
\begin{bmatrix}
\cos\psi & -\sin\psi & 0\\
\sin\psi & \cos\psi  & 0\\
0        & 0         & 1
\end{bmatrix},
\end{equation}
$M\in\mathbb{R}^{3\times 3}$ is the effective inertia matrix, 
$D\in\mathbb{R}^{3\times 3}$ is the damping matrix, and 
$\tau_c\in\mathbb{R}^3$ is the control wrench generated by the pushing 
agents.\subsection{Contact geometry}
\begin{figure}[!t]
    \centering
    \includegraphics[width=0.75\linewidth]{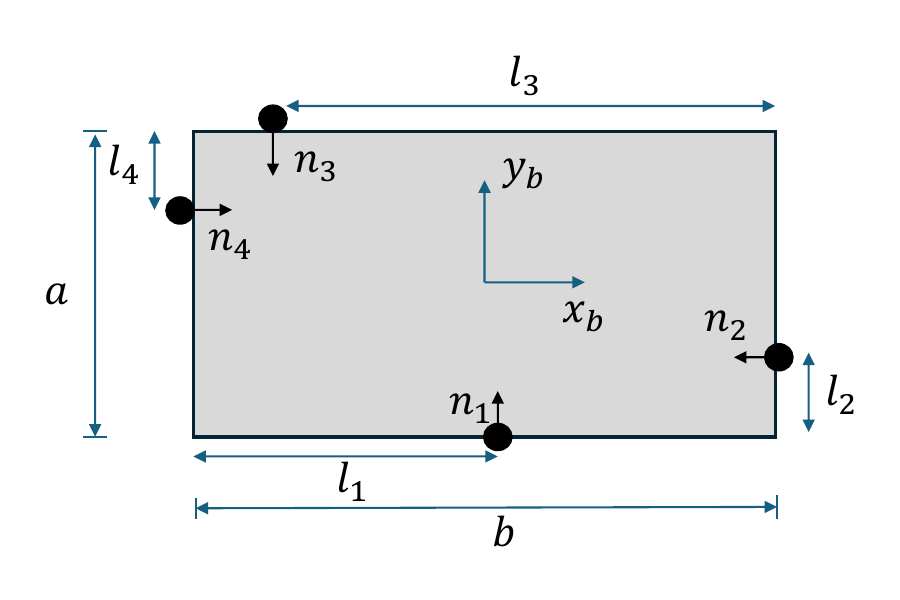}
    \caption{Contact geometry of the rectangular object. A body-fixed frame is attached to the geometric center. The four contact points are parameterized by $l_1,l_2,l_3,l_4$ along the object boundary, and the corresponding inward unit normals are denoted by $n_1,n_2,n_3,n_4$.}
    \label{fig_geometry}
\end{figure}

Assume that four agents apply unilateral frictionless pushing forces at four points on the rectangle boundary, one on each edge. Let the rectangle length and width be denoted by $b>0$ and $a>0$, respectively. A body-fixed frame is attached to the geometric center of the rectangle. The four contact locations are parameterized by
\begin{equation}
l_1,l_3\in(0,b),\qquad l_2,l_4\in(0,a),
\end{equation}
where $l_1$ is measured along the bottom edge from the left corner, $l_2$ along the right edge from the bottom corner, $l_3$ along the top edge from the right corner, and $l_4$ along the left edge from the top corner, as shown in Fig.~\ref{fig_geometry}.

Under this convention, the contact points expressed in the body frame are
\begin{align}
p_1 &= \begin{bmatrix} -\frac{b}{2}+l_1 \\ -\frac{a}{2} \end{bmatrix},&
p_2 &= \begin{bmatrix} \frac{b}{2} \\ -\frac{a}{2}+l_2 \end{bmatrix}, \nonumber\\
p_3 &= \begin{bmatrix} \frac{b}{2}-l_3 \\ \frac{a}{2} \end{bmatrix},&
p_4 &= \begin{bmatrix} -\frac{b}{2} \\ \frac{a}{2}-l_4 \end{bmatrix}.
\end{align}

The corresponding inward unit normals are
\begin{equation}
n_1=\begin{bmatrix}0\\1\end{bmatrix},\quad
n_2=\begin{bmatrix}-1\\0\end{bmatrix},\quad
n_3=\begin{bmatrix}0\\-1\end{bmatrix},\quad
n_4=\begin{bmatrix}1\\0\end{bmatrix}.
\end{equation}

Each agent applies a unilateral force of the form
\begin{equation}
f_i = \lambda_i n_i,\qquad \lambda_i \ge 0,\qquad i=1,\dots,4.
\end{equation}
Let
\begin{equation}
\lambda =
\begin{bmatrix}
\lambda_1 & \lambda_2 & \lambda_3 & \lambda_4
\end{bmatrix}^\top
\end{equation}
be the vector of nonnegative contact-force magnitudes. The corresponding planar contact wrench is
\begin{equation}
w_i =
\begin{bmatrix}
n_i\\
p_i \times n_i
\end{bmatrix},
\qquad i=1,\dots,4,
\end{equation}
where $p_i \times n_i = p_{ix}n_{iy} - p_{iy}n_{ix}$ denotes the scalar moment about the geometric-center frame. Let
\begin{equation}
W=\{w_1,w_2,w_3,w_4\}
\end{equation}
denote the set of generating planar wrenches. The total planar wrench generated by the four contacts is
\begin{equation}
w=\sum_{i=1}^4 \lambda_i w_i.
\end{equation}
For compact matrix notation, substituting the contact points and inward normals into the wrench definition gives the contact wrench matrix
\begin{equation}
\begin{aligned}
G
&=
\begin{bmatrix}
w_1 & w_2 & w_3 & w_4
\end{bmatrix} \\
&=
\begin{bmatrix}
0 & -1 & 0 & 1\\
1 & 0 & -1 & 0\\
l_1-\frac{b}{2} & l_2-\frac{a}{2} & l_3-\frac{b}{2} & l_4-\frac{a}{2}
\end{bmatrix}.
\end{aligned}
\label{eq_wrench_matrix}
\end{equation}
Hence,
\begin{equation}
w=G\lambda.
\end{equation}

\subsection{Problem Statement}
Consider a rectangular floating object subject to four unilateral frictionless pushing contacts, with one contact located on each edge. The contact locations are parameterized by $(l_1,l_2,l_3,l_4)$, as shown in Fig.~\ref{fig_geometry}. Since each contact can only push along the inward normal of the corresponding edge, the set of wrenches generated by the contacts depends directly on these four locations.

The problem addressed in this paper is to determine the necessary and sufficient condition on $(l_1,l_2,l_3,l_4)$ under which the four-contact configuration can generate arbitrary planar wrenches on the rectangular object. In the frictionless planar setting considered here, this is equivalent to determining when the contact configuration achieves force closure.

This geometric condition is independent of both the desired trajectory and the center of mass of the object. In the control framework, it is used as a contact-feasibility check before contact force allocation, ensuring that the current contact configuration is capable of producing the planar wrench required for trajectory tracking.

\section{CLOSED-FORM FORCE-CLOSURE CONDITION}

This section derives an explicit necessary and sufficient condition for four unilateral edge-normal contacts to generate arbitrary planar wrenches on the rectangular object.

\subsection{Force-closure condition}

Under the unilateral frictionless pushing model, the wrench cone generated by the contacts is
\begin{equation}
\operatorname{cone}(W) = \left\{ G\lambda \;:\; \lambda \ge 0 \right\}.
\end{equation}
Force closure is achieved if and only if $\operatorname{cone}(W)=\mathbb{R}^3$, that is, any planar wrench can be generated by nonnegative contact forces.

Before deriving the closed-form condition, we first state the minimum contact placement requirement. For planar frictionless point contact models, four unilateral contacts are the standard minimum set for force closure of a general planar object~\cite{nguyen1988constructing}.

\begin{proposition}
For a rectangular object with unilateral frictionless edge-normal contacts, force closure is impossible unless at least one contact is placed on each edge.
\end{proposition}

\begin{proof}
Force closure requires the wrench cone to span $\mathbb{R}^3$; hence its translational projection must span $\mathbb{R}^2$. With frictionless pushing, the only possible force directions are the four inward normals. Because contacts are unilateral, each direction appears with a nonnegative coefficient only. If any edge has no contact, the corresponding direction is absent, so the translational cone cannot equal $\mathbb{R}^2$ and force closure fails.
\end{proof}

\begin{theorem}
Consider the configuration in Fig.~\ref{fig_geometry}, where one unilateral frictionless contact is placed on each edge. The contact configuration achieves force closure if and only if
\begin{equation}
(l_1+l_3-b)(l_2+l_4-a)<0.
\label{eq_main_condition}
\end{equation}
\end{theorem}

\begin{proof}
By the standard planar frictionless force-closure characterization~\cite{nguyen1988constructing}, force closure is equivalent to the existence of $\lambda > 0$ satisfying $G\lambda = 0$, together with $\mathrm{rank}(G)=3$.

\textit{Necessity.} Assume force closure holds. Then there exists $\lambda_i > 0$ such that $G\lambda = 0$. From \eqref{eq_wrench_matrix}, the first two rows give $\lambda_1 = \lambda_3 = \alpha > 0$ and $\lambda_2 = \lambda_4 = \beta > 0$. The third row then requires
\begin{equation}
\alpha(l_1+l_3-b)+\beta(l_2+l_4-a)=0.
\label{eq_moment_reduced}
\end{equation}
Since $\alpha, \beta > 0$, the two coefficients must have opposite signs, so $(l_1+l_3-b)(l_2+l_4-a)<0$.

\textit{Sufficiency.} Assume \eqref{eq_main_condition} holds. Then $l_1+l_3-b$ and $l_2+l_4-a$ have opposite signs, so there exist $\alpha, \beta > 0$ satisfying \eqref{eq_moment_reduced}. Setting $\lambda_1 = \lambda_3 = \alpha$, $\lambda_2 = \lambda_4 = \beta$ gives $\lambda > 0$ and $G\lambda = 0$. For rank, the $3 \times 3$ minors for column sets $\{1,2,3\}$ and $\{1,2,4\}$ have determinants
\begin{equation}
\det(G_{123})=l_1+l_3-b, \qquad \det(G_{124})=l_2+l_4-a.
\end{equation}
Under \eqref{eq_main_condition}, both are nonzero, so $\mathrm{rank}(G)=3$.
\end{proof}

\begin{remark}
The condition $(l_1+l_3-b)(l_2+l_4-a)<0$ has a direct geometric interpretation in terms of pairwise moment balance. For each opposing normal pair, the two contact points define a signed relative offset along the corresponding edge direction. Force closure requires the two opposing pairs to have relative offsets with opposite signs. This opposite moment arm arrangement allows the unilateral contacts to generate planar wrenches in all directions.
\end{remark}
\subsection{Independence from the center-of-mass location}

The condition in \eqref{eq_main_condition} is derived with the wrench reference point at the geometric center. This subsection shows that force closure is invariant to the choice of reference point, so the condition holds regardless of the center-of-mass location.

\begin{figure*}[!t]
    \centering
     \includegraphics[
        width=0.75\linewidth
      ]{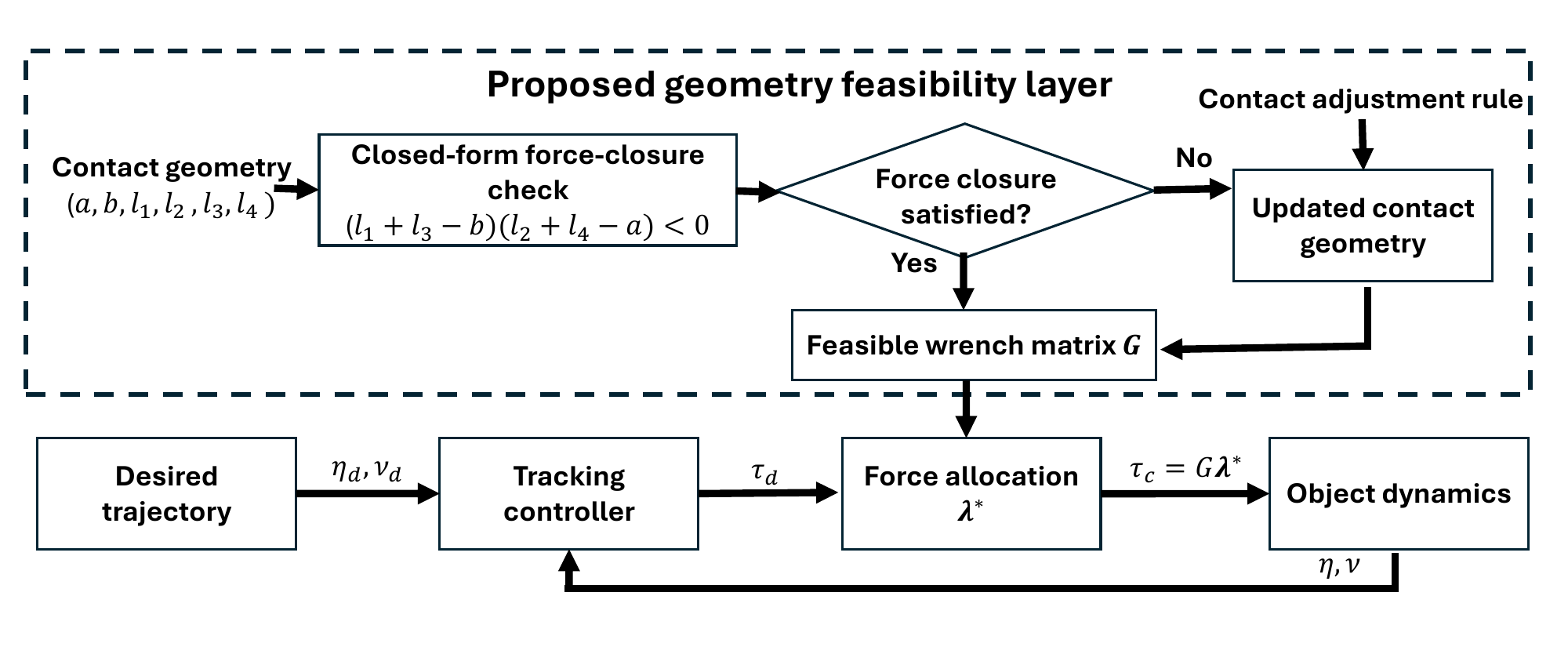}
    \caption{Integration of the proposed geometric feasibility layer into a conventional wrench-based control framework. The contact geometry is checked using the proposed closed-form force closure condition before contact force allocation. If the condition is not satisfied, the contact adjustment rule updates the configuration to obtain a feasible wrench matrix $G$. The tracking controller generates the commanded wrench $\tau_d$, which is allocated to nonnegative pushing forces $\lambda^\ast$ for object motion control.}
    \label{fig_framework}
\end{figure*}

\begin{proposition}
Let the wrench reference point be shifted by an arbitrary vector $d=\begin{bmatrix} d_x & d_y \end{bmatrix}^\top\in\mathbb{R}^2$. Then the force-closure condition in \eqref{eq_main_condition} remains unchanged. Consequently, it is independent of the object center-of-mass location.
\end{proposition}

\begin{proof}
Under a shift by $d$, each planar wrench transforms as $w_i' = T(d)w_i$, where
\begin{equation}
T(d)=
\begin{bmatrix}
1 & 0 & 0\\
0 & 1 & 0\\
d_y & -d_x & 1
\end{bmatrix}.
\end{equation}
Since $T(d)$ is invertible for any $d\in\mathbb{R}^2$, the transformed wrench cone satisfies $\operatorname{cone}(W') = T(d)\operatorname{cone}(W)$. Therefore,
\begin{equation}
\operatorname{cone}(W)=\mathbb{R}^3
\iff
\operatorname{cone}(W')=\mathbb{R}^3,
\end{equation}
and force closure is invariant to the choice of wrench reference point.
\end{proof}

\section{GEOMETRIC FEASIBILITY LAYER FOR CONTACT FORCE ALLOCATION}
Since the proposed force-closure condition is independent of the task and controller, it can be used as a contact-feasibility check before contact force allocation. This section integrates the condition into a simple control framework for cooperative pushing of a floating rectangular object, as shown in Fig.~\ref{fig_framework}. The framework first checks the contact geometry, corrects it if necessary, and then performs contact force allocation and trajectory tracking.

\subsection{Contact adjustment rule}
When the force-closure condition is not satisfied, the contact configuration must be corrected before contact force allocation can proceed. Let $A = l_1 + l_3 - b$ and $B = l_2 + l_4 - a$. Force closure requires $AB < 0$, i.e., $A$ and $B$ must have opposite signs. If $AB \geq 0$, the correction identifies whichever of 
$|A|$ or $|B|$ is smaller and shifts the corresponding contact 
pair so that its sign is reversed. For instance, if $|A| \leq |B|$, the bottom and top contacts are repositioned so that 
$l_1^+ + l_3^+ = b - \delta\,\mathrm{sign}(B)$, where $\delta > 0$ is a small margin. This makes the corrected values of $A$ and $B$ have opposite signs, thereby restoring force closure. This rule is not an optimal planner, but a simple way to restore wrench feasibility before contact force allocation.

\subsection{Force allocation and trajectory tracking}

Once force closure is confirmed, contact force allocation and tracking control proceed. The contact forces are computed using a nonnegative least-squares allocation:
\begin{equation}
\lambda^\star =
\arg\min_{\lambda\geq 0}
\left[
\frac{1}{2}\|S(G\lambda-\tau_d)\|_2^2
+
\frac{\rho}{2}\|\lambda\|_2^2
\right],
\end{equation}
where $S$ is a diagonal scaling matrix, $\rho>0$ is a small regularization weight, and $\tau_d$ is the commanded wrench. The applied wrench is $\tau_c = G\lambda^\star$. A force-closure configuration can match arbitrary wrench commands using nonnegative pushing forces, whereas a non-force-closure configuration can only return the closest wrench inside its unilateral wrench cone.
For the tracking controller, let the desired trajectory be 
$\eta_d(t)=[x_d,\,y_d,\,\psi_d]^{\top}$ with body-frame velocity 
$\nu_d(t)=[u_d,\,v_d,\,r_d]^{\top}$. The pose error in the body frame is
\begin{equation}
e_{\eta}=
\begin{bmatrix}
R^{\top}(\psi)(p_d-p)\\
\psi_d-\psi
\end{bmatrix},
\end{equation}
where $p=[x\;y]^\top$, $p_d=[x_d\;y_d]^\top$, 
$R(\psi)$ denotes the planar rotation matrix, mapping body-frame vectors to the world frame,
and the velocity error is $e_{\nu}=\nu_d-\nu$. The commanded wrench is
\begin{equation}
\tau_d = K_p e_{\eta} + K_d e_{\nu},
\end{equation}
where $K_p,K_d\in\mathbb{R}^{3\times 3}$ are positive definite gain matrices.
\section{SIMULATION RESULTS AND DISCUSSION}

This section presents two simulation studies. The first study verifies whether each contact configuration can generate representative commanded planar wrenches under unilateral inward-normal pushing forces. The second study evaluates full-pose trajectory tracking through a narrow passage, where the object must follow both planar translation and heading commands.

\subsection{Simulation setup}
\begin{figure}[!b]
    \centering
    \includegraphics[width=0.74\linewidth]{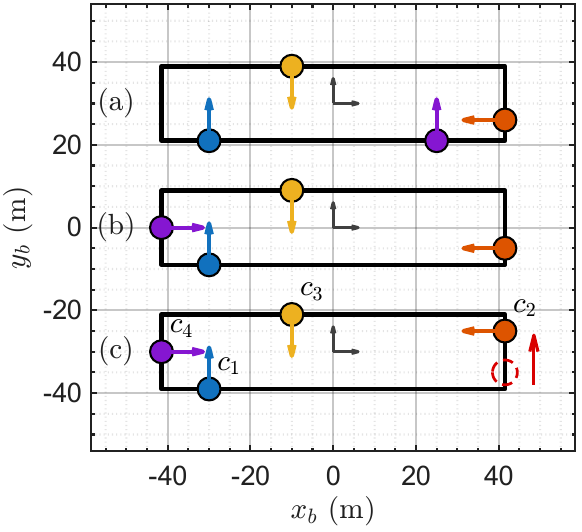}
   \caption{Contact configuration scenarios: (a) missing edge coverage, (b) one contact per edge without force closure, and (c) one contact per edge with force closure after moving $c_2$.}
    \label{fig_part1_contact_scenarios}
\end{figure}

The simulations use a reduced 3-DOF planar floating-body model to isolate the effect of contact geometry on wrench generation. The surge-sway-yaw inertia matrix is taken from the offshore supply vessel (OSV) model in the Marine Systems Simulator (MSS)~\cite{fossen2011handbook,perez2006overview}, with $M=10^6[5.58,0,0;0,6.98,-31.5;0,-31.5,3490]$ and $D=10^6\mathrm{diag}(0.12,\ 0.18,\ 75.0)$. The body is modeled as an equivalent rectangle with length $b=83$ m and width $a=18$ m. The pushing robots are modeled as ideal unilateral actuators that generate normal forces at the contact points. A PD trajectory tracking controller generates a commanded planar wrench $\tau_d$ from the pose and velocity tracking errors. The contact force allocation then computes unilateral contact forces by minimizing the wrench error subject to nonnegative force constraints. Table~\ref{table_contact_configurations} and Fig.~\ref{fig_part1_contact_scenarios} summarize the three contact configurations. Scenario 1 lacks full edge coverage, so $\Phi$ is not evaluated. Scenario 2 has $A<0$ and $B<0$, giving $\Phi>0$ and no force closure. Since $|B|<|A|$, moving $c_2$ changes $l_2$ from $4.0$ m to $14.0$ m, making $B>0$. Scenario 3, therefore, has $\Phi<0$ and satisfies the force-closure condition.

\begin{table}[h]
\caption{Contact configurations and force-closure condition values, where $A=l_1+l_3-b$, $B=l_2+l_4-a$, and $\Phi=AB$.}
\label{table_contact_configurations}
\centering
\setlength{\tabcolsep}{3.2pt}
\renewcommand{\arraystretch}{1.12}
\begin{tabular}{c c c c c c c c c}
\toprule
Scenario & $l_1(m)$ & $l_2(m)$ & $l_3(m)$ & $l_4(m)$ & $A$ & $B$ & $\Phi$ & FC \\
\midrule
1 & 11.5, 66.5 & 5.0 & 51.5 & N/A & N/A & N/A & N/A & No \\
2 & 11.5 & 4.0 & 51.5 & 9.0 & $-20$ & $-5$ & 100 & No \\
3 & 11.5 & 14.0 & 51.5 & 9.0 & $-20$ & 5 & $-100$ & Yes \\
\bottomrule
\end{tabular}
\vspace{-5mm}
\end{table}
\subsection{Static planar-wrench feasibility}

The wrench commands are $\tau_d=[10,8,0]^\top$, $[0,0,120]^\top$, $[0,0,-120]^\top$, and $[10,8,120]^\top$, with forces in $kN$ and yaw moment in $kNm$. These commands test planar force generation, positive and negative yaw moments, and coupled force-moment generation.

Table~\ref{table_static_wrench_feasibility} summarizes the errors between the commanded and realized wrenches, along with the feasibility results. The results are consistent with the proposed condition. Scenario 1 cannot guarantee arbitrary planar wrench generation because one edge direction is missing, consistent with Proposition 1. Scenario 2, although using one contact on each edge, violates the proposed condition and can realize only a restricted subset of commanded wrenches. In contrast, Scenario 3 satisfies the force-closure condition and produces negligible residuals for all tested wrench commands. This confirms that the proposed condition correctly determines whether full planar wrench generation is achievable.

\begin{table}[!t]
\vspace{0.2em}
\caption{Static planar-wrench feasibility results}
\label{table_static_wrench_feasibility}
\centering
\setlength{\tabcolsep}{4pt}
\renewcommand{\arraystretch}{1.12}
\begin{tabular}{c c c c c c c}
\toprule
Scenario & Case & FC & $e_{F_x}$ & $e_{F_y}$ & $e_{M_z}$ & Feasible \\
 & $\tau_d=[F_x,F_y,M_z]^\top$  &  & (kN) & (kN) & (kNm) &  \\
\midrule
\multirow{4}{*}{1}
& $[10,8,0]^\top$     & No  & 10.0 & 0    & 0    & No  \\
& $[0,0,120]^\top$    & No  & 0    & 0    & 0    & Yes \\
& $[0,0,-120]^\top$   & No  & 0    & 0    & 0    & Yes \\
& $[10,8,120]^\top$   & No  & 10.0 & 0    & 0    & No  \\
\midrule
\multirow{4}{*}{2}
& $[10,8,0]^\top$     & No  & 0    & 8.0  & 0.3  & No  \\
& $[0,0,120]^\top$    & No  & 0    & 11.9 & 1.2  & No  \\
& $[0,0,-120]^\top$   & No  & 0    & 0    & 0    & Yes \\
& $[10,8,120]^\top$   & No  & 0    & 19.8 & 2.0  & No  \\
\midrule
\multirow{4}{*}{3}
& $[10,8,0]^\top$     & Yes & 0    & 0    & 0    & Yes \\
& $[0,0,120]^\top$    & Yes & 0    & 0    & 0    & Yes \\
& $[0,0,-120]^\top$   & Yes & 0    & 0    & 0    & Yes \\
& $[10,8,120]^\top$   & Yes & 0    & 0    & 0    & Yes \\
\bottomrule
\end{tabular}
\vspace{-2.5mm}
\end{table}

\subsection{Full-pose tracking through a narrow passage}
The second study examines the effect of the proposed force-closure condition on dynamic full-pose tracking. Scenarios 2 and 3 are compared in a narrow-passage maneuver using the same controller and contact force allocation. The object must track both position and yaw while passing through a constrained region, requiring coupled force and moment generation rather than pure translation.

\begin{figure}[!t]
    \centering
    \vspace{0.6em}
    \includegraphics[
        width=0.8\linewidth,
        trim=0 0 0 0.42cm,
        clip
    ]{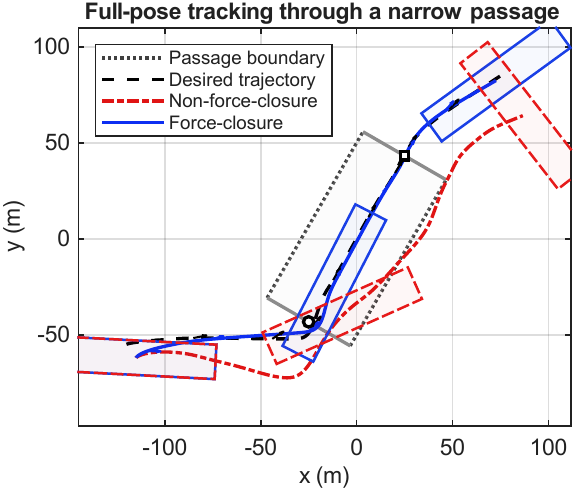}
    \caption{Top-view narrow-passage tracking comparison showing path deviation and boundary violation without force closure.}
    \label{fig_narrow_gate_topview}
 
\end{figure}

\begin{figure}[!t]
    \centering
    \scalebox{1}[1.25]{%
        \includegraphics[width=1\linewidth]{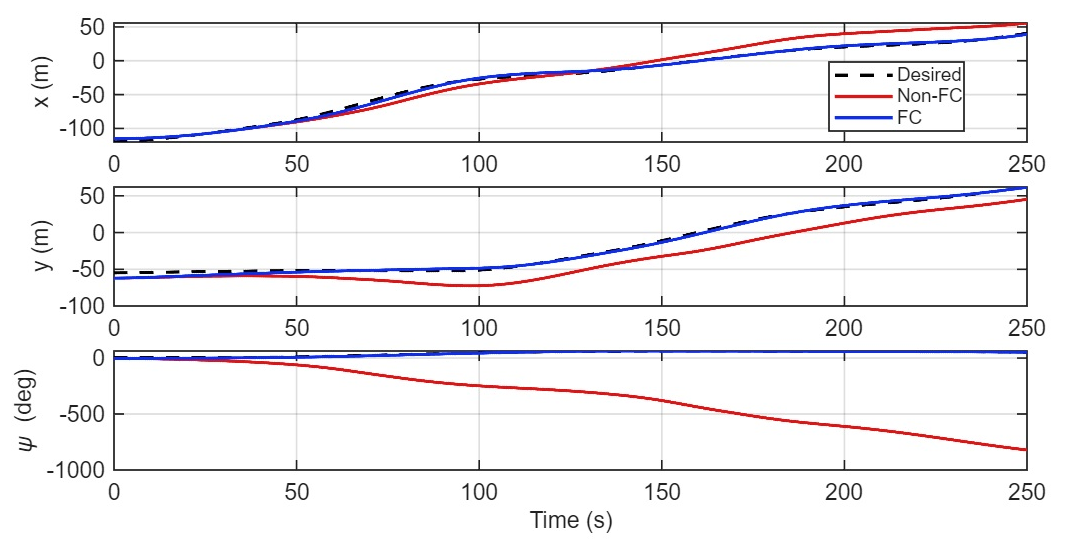}
    }
    \caption{Time-domain tracking of $x$, $y$, and $\psi$, showing accumulated yaw error when force closure is not satisfied.}
    \label{fig_pose_tracking}
\end{figure}

As shown in Fig.~\ref{fig_narrow_gate_topview}, the force-closure configuration in Scenario 3 follows the desired path and passes through the narrow passage, while the configuration without force closure in Scenario 2 cannot maintain the required yaw angle, causing the object trajectory to deviate and violate the passage boundary. The time domain results in Fig.~\ref{fig_pose_tracking} further show that this yaw error accumulates over time: Scenario 3 tracks the desired $x$, $y$, and $\psi$ profiles with RMS errors of $2.24$ m, $2.66$ m, and $2.69^\circ$, whereas Scenario 2 fails with RMS errors of $12.05$ m, $18.11$ m, and $535.96^\circ$ in $x$, $y$, and unwrapped yaw, respectively. These results show that the proposed condition affects full-pose tracking performance, not only static wrench feasibility.


\begin{figure}[!t]
    \centering
    \includegraphics[width=1.02\linewidth]{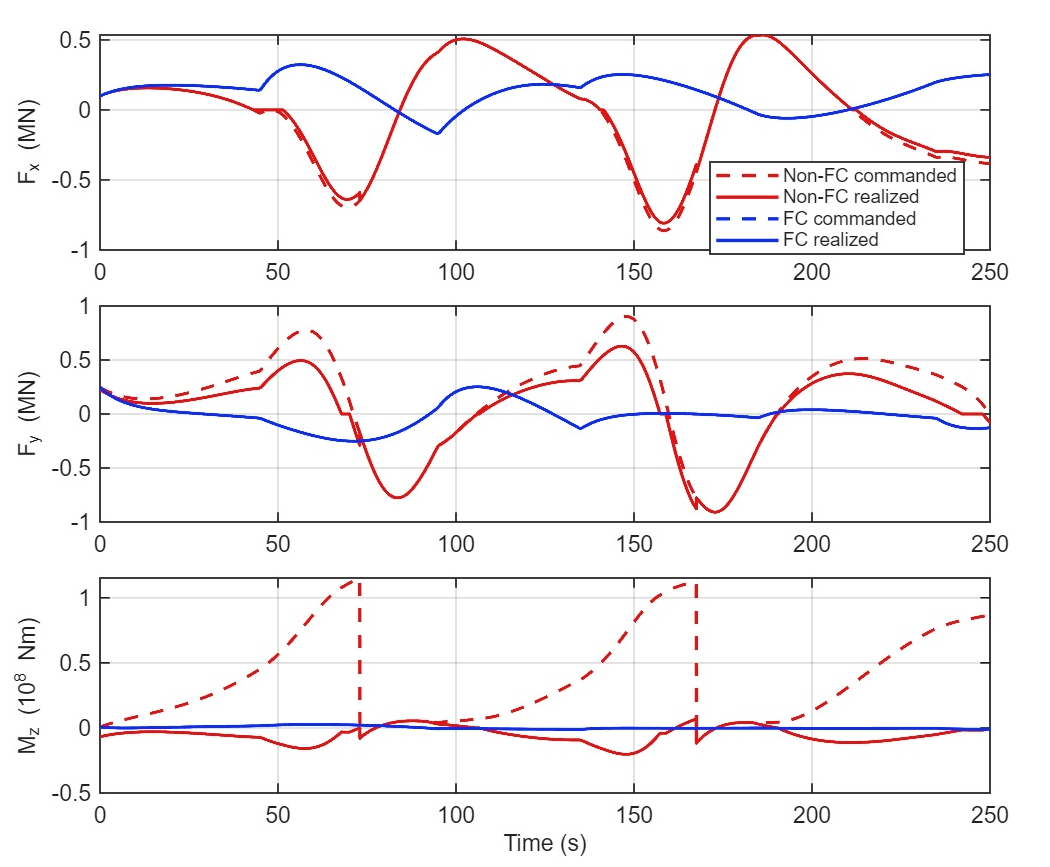}
   \caption{Commanded and realized wrenches during the narrow-passage maneuver with the force-closure and non-force-closure configurations.}
    \label{fig_control_wrench}
    \vspace{-3mm}
\end{figure}
The wrench profiles in Fig.~\ref{fig_control_wrench} explain the performance gap between the two configurations. In Scenario 3, the force-closure configuration reproduces the commanded wrench closely, while in Scenario 2, not all wrench commands can be generated, leading to the observed position and yaw tracking errors. In this floating body model, the planar dynamics include sway-yaw coupling terms in the inertia matrix, and the pose kinematics further couple heading with translational motion. Therefore, a mismatch between the commanded and realized yaw moment can contribute to both orientation and position tracking errors. This highlights the value of integrating the proposed force-closure condition within a planar floating body control framework.

\section{CONCLUSION}
This paper derived a necessary and sufficient closed-form condition for planar force closure of a rectangular object under four frictionless unilateral pushing contacts. The condition depends only on the four edge contact positions and can be used as a geometric feasibility layer before contact force allocation. When the condition is not satisfied, the signs of its factors identify which opposing contact pair should be adjusted toward force closure. Static wrench tests and floating-body tracking simulations showed that force-closure configurations reproduce the commanded planar wrenches with smaller tracking errors, while configurations without force closure lead to wrench errors and poor tracking. Future work could extend the framework to more realistic marine settings by incorporating hydrodynamic disturbances, autonomous tugboat dynamics, and more general object geometries.

\bibliographystyle{IEEEtran}
\bibliography{Reference}

\end{document}